\documentclass[runningheads]{llncs}

\usepackage{pgfplots}
\pgfplotsset{compat=1.17}
\usepackage{graphicx}
\usepackage{subcaption}
\usepackage{adjustbox}
\usepackage{booktabs}
\usepackage{amsmath}
\usepackage{amssymb}
\usepackage[hidelinks]{hyperref}
\usepackage[capitalize,nameinlink]{cleveref}
\usepackage[nocompress]{cite}
\usepackage[misc,geometry]{ifsym}

\usepackage{tikz}
\usetikzlibrary{positioning,calc}

\usepackage[color=black,backgroundcolor=white]{todonotes}

\begin{document}
\title{Self-Supervised Vision Transformers for Writer Retrieval}
%
\author{Tim Raven\textsuperscript{~\Letter~} \orcidID{0009-0002-1528-2744}   \and
Arthur Matei \orcidID{0009-0009-6028-7502}       \and
Gernot A. Fink  \orcidID{0000-0002-7446-7813}      }

\authorrunning{T. Raven\textit{ et al.}}

\institute{TU Dortmund University \\
\email{{\{tim.raven, arthur.matei, gernot.fink\}@tu-dortmund.de}}\\}

\maketitle              
\begin{abstract}

While methods based on Vision Transformers (ViT) have achieved state-of-the-art performance in many domains, they have not yet been applied successfully in the domain of writer retrieval. The field is dominated by methods using handcrafted features or features extracted from Convolutional Neural Networks.  In this work, we bridge this gap and present a novel method that extracts features from a ViT and aggregates them using VLAD encoding. The model is trained in a self-supervised fashion without any need for labels. We show that extracting local foreground features is superior to using the ViT's class token in the context of writer retrieval. We evaluate our method on two historical document collections. We set a new state-at-of-art performance on the Historical-WI dataset (83.1\% mAP), and the HisIR19 dataset (95.0\% mAP). Additionally, we demonstrate that our ViT feature extractor can be directly applied to modern datasets such as the CVL database (98.6\% mAP) without any fine-tuning.

\end{abstract}
\section{Introduction}
Writer retrieval involves systematically extracting documents, written by the same author as a queried document, from a large corpus of handwritten texts with unidentified authors. Closely related to this, writer identification aims to identify the writer of a queried document by consulting a corpus of labeled documents. In historical research, these processes are crucial for categorizing and examining manuscripts based on authorship, particularly when manuscripts lack signatures~\cite{christlein_icdar_2019}. In the forensic sciences, accurately identifying authors in handwritten notes or documents is essential in criminal investigations, such as verifying ransom notes, anonymous threatening letters, or fraudulent documents.

Recently, methods employing Vision Transformers (ViT)~\cite{dosovitskiy_image_2021} have achieved state-of-the-art performance in various computer vision tasks, including handwritten text recognition ~\cite{li_trocr_2022}.
However, ViTs are prone to overfitting even on large datasets, making them challenging to train~\cite{dosovitskiy_image_2021}.
Self-supervised learning can help to address this challenge, as no annotations are required for training and, even when annotations are available, self-supervised training followed by supervised fine-tuning can outperform fully supervised training~\cite{he_masked_2022}. 
Popular self-supervised learning paradigms include contrastive learning~\cite{chen_simple_2020,chen_empirical_2021}, self-distillation~\cite{caron_unsupervised_2021}, masked image modeling~\cite{bao_beit_2022,he_masked_2022,xie_simmim_2022} and similarity maximization~\cite{chen_exploring_2021}; or a combination of these~\cite{zhou_ibot_2022,kakogeorgiou_what_2022}. 

Current writer retrieval and identification methods still rely on local features extracted from Convolutional Neural Network (CNN) activations~\cite{christlein_unsupervised_2017} or handcrafted methods~\cite{lai_encoding_2020}. While there have been efforts to employ ViTs for writer retrieval~\cite{peer_self-supervised_2022}, the results were not competitive.  
Motivated by the recent success of pre-training a ViT for handwritten text recognition using masked image modeling~\cite{souibgui_text-diae_2022} on only IAM~\cite{marti2002iam}, we revisit the use of self-supervised ViTs in writer retrieval.

\begin{figure}[t]
\centering
\includegraphics[width=\textwidth]{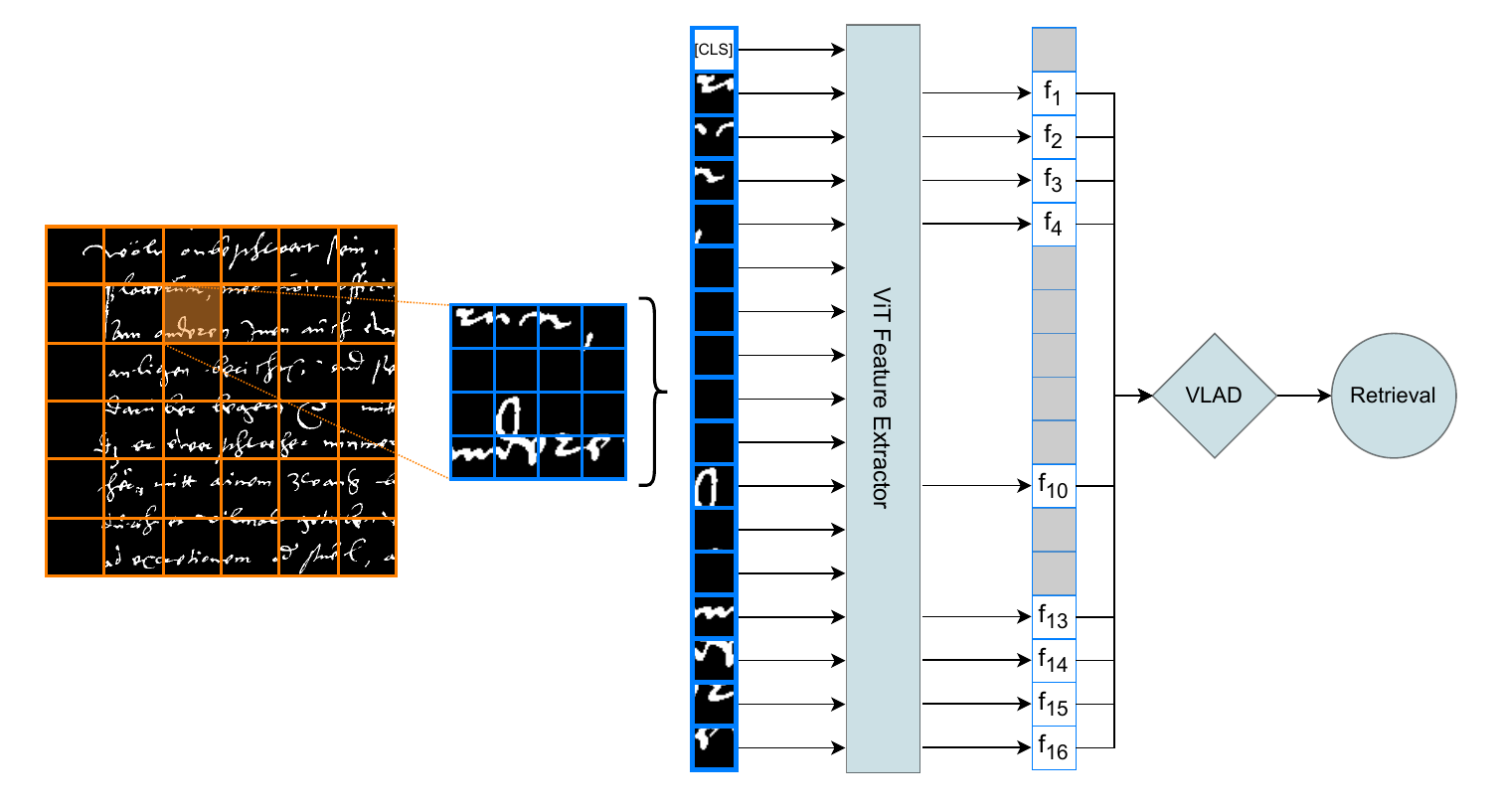}

\caption{Illustration of our proposed method. Document images are cut into windows in a regular grid. The windows are again cut into patches and form the input sequence. To extract local features, a self-supervised Vision Transformer (ViT) is used. We extract only foreground patch tokens from the ViT output sequence, i.e., only the patch tokens of input patches with sufficient handwriting. All foreground tokens of the document are aggregated using a VLAD encoding. These encodings are used for retrieval and reranking.}
\label{fig:overview}
\end{figure}

In this work, we introduce a fully self-supervised approach that employs a ViT as a feature extractor for writer retrieval and identification (see \cref{fig:overview}). Our method requires no labels and outperforms existing methods in historical document benchmarks. We train the ViT using a combination of Masked Image Modeling \cite{he_masked_2022,xie_simmim_2022} and self-distillation~\cite{caron_unsupervised_2021}, utilizing a student-teacher network. 
The teacher's self-attention is used to guide the masking process which in turn the student has to reconstruct in feature space. 
Contrary to previous attempts of utilizing a ViT as a feature extractor for writer retrieval\cite{peer_self-supervised_2022}, we do not use the class token of the ViT as feature representation. Instead, we extract foreground patch tokens from the ViT's output sequence, i.e., tokens corresponding to input patches with a sufficient amount of foreground pixels. Our evaluation demonstrates that encoding these features with a Vector of Locally Aggregated Descriptors (VLAD)~\cite{jegou_aggregating_2011} to obtain a global page descriptor enhances performance notably compared to using the class token with either sum-pooling or VLAD. However, even when using sum-pooling as encoding, our method surpasses previous methods, underscoring the quality of the extracted features.

We make the following contributions:
\begin{itemize}
    \item We successfully apply a Vision Transformer to the task of writer retrieval through self-supervised learning.
    \item We demonstrate that features extracted from our ViT outperform both handcrafted and CNN-based features.
    \item We show that encoding foreground tokens using VLAD is superior to encoding class tokens.
    \item We show that our method learns robust features directly applicable to historical and modern handwriting without the need for model fine-tuning.
\end{itemize}

The remainder of this paper is organized as follows. First, we cover related work in the field of writer retrieval in \cref{related}. Next, we outline our proposed method in more detail in \cref{method}. Then, we introduce the evaluation protocol for our experiments in \cref{eval}. The conducted experiments are detailed in \cref{exp}. Finally, we summarize our findings and give an outlook on future work in \cref{conc}. 

%

\section{Related Work} \label{related}
Writer retrieval methods commonly follow the same pipeline of local feature extraction, page-level aggregation, and distance-based retrieval with optional reranking. 

\subsection{Feature Extraction} Local features are generally divided into handcrafted features and deep learning features. Examples of handcrafted features are SURF~\cite{jain_combining_2014}, Zernike moments~\cite{christlein_writer_2015} or a combination of SIFT and Pathlet features~\cite{lai_encoding_2020}. In contrast, deep learning features were first introduced by Fiel and Sablatnig~\cite{fiel_writer_2015}, who used a supervised CNN as a feature extractor. Christlein\textit{ et al.}~\cite{christlein_unsupervised_2017} propose an unsupervised method for CNN training for historical documents, in which pseudo labels are derived from clustering SIFT descriptors. 
In~\cite{peer_self-supervised_2022} an ImageNet pre-trained ViT is fine-tuned for document images in a self-supervised student-teacher approach operating on differently augmented views, using the ViT's class tokens as feature representation. However, the method underperforms considerably compared to CNN-based or handcrafted methods. 

\subsection{Aggregation} Methods to aggregate local features into global page descriptors are generally categorized into codebook-free and codebook-based methods. Codebook-free methods include sum-pooling and generalized max-pooling~\cite{murray2014generalized}. Codebook-based methods used in writer retrieval/identification include Fisher Vectors~\cite{fiel_writer_2013}, GMM supervectors~\cite{christlein_writer_2014} and VLAD-based methods~\cite{christlein_unsupervised_2017,lai_encoding_2020}, where several VLAD encodings are computed and jointly decorrelated using PCA with whitening. The VLAD encoding is sometimes incorporated into the network architecture using NetVLAD~\cite{rasoulzadeh_writer_2022,peer_towards_2023}. As an additional refinement step, Christlein\textit{ et al.}~\cite{christlein_unsupervised_2017} train exemplar SVMs for each query, using the query as the only positive example and all training pages as negatives, exploiting the writer-disjointness of training and test sets. We find that a single VLAD encoding is sufficient with our features.

\subsection{Retrieval and Reranking} Retrieval is commonly done using a distance measure, like cosine distance. Other metrics such as the $l_1$ distance, $l_2$ distance, or the Canberra distance have also been explored \cite{peer2022distances}. An optional step that has shown to be beneficial in retrieval tasks is reranking, which refines the retrieval list by exploiting the information within it. The authors of~\cite{rasoulzadeh_writer_2022} use a $k$ reciprocal nearest neighbor Query Expansion by averaging each descriptor with its $k$ reciprocal nearest neighbors. The E-SVM approach of Christlein \textit{et al.} is extended into a Pair/Triple SVM approach in~\cite{jordan_re-ranking_2020}, where similar documents from the test set are included as additional \textit{positive} examples. In~\cite{peer_towards_2023} (Similarity) Graph Reranking is explored. Here, a parameter-free graph network is constructed and used to obtain updated global descriptors using message propagation.

\section{Method} \label{method}

 Our method follows the common framework of local feature extraction, aggregation, distance-based retrieval and reranking used in previous work~\cite{christlein_unsupervised_2017,christlein_encoding_2018,peer_towards_2023,lai_encoding_2020}. 
 Our method operates on binary images, obtained in a preprocessing step, if necessary. 
 An illustration of the method is given in \Cref{fig:overview}. As the computational complexity of a ViT grows quadratically with the sequence length, our ViT uses a fixed input image size of $224 \times 224$, and operates on windows extracted from the document.
 The ViT is trained in a self-supervised fashion (see \Cref{meth:pretrain}) and used to extract patch tokens corresponding to handwriting (see \Cref{meth:extract}). The extracted features are aggregated into a global page descriptor using a VLAD encoding (see \Cref{meth:aggregate}). Finally, the global page descriptors are compared using cosine distance and optionally reranked (see \Cref{meth:rerank}).

\subsection{Self-supervised Training} \label{meth:pretrain} 


ViTs feature a large number of trainable parameters and are prone to overfit \cite{dosovitskiy_image_2021}, requiring extensive amounts of annotated data to train a ViT successfully with traditional, supervised methods. 
Thus, utilizing self-supervised training for ViTs is a logical conclusion. 
We chose AttMask~\cite{kakogeorgiou_what_2022} for self-supervised training. The method is an adaptation of iBOT~\cite{zhou_ibot_2022}, which incorporates Masked Image Modeling (MIM) into the DINO~\cite{caron_unsupervised_2021} framework.

DINO~\cite{caron_unsupervised_2021} employs self-distillation, a special form of knowledge distillation, i.e., training a student network to reproduce the output of a teacher network. In self-distillation, the teacher network is defined as an exponential moving average of the student.
Student and teacher receive differently augmented views of the input, forcing the student to learn an invariance to the applied augmentations. Additionally, the method samples global and local views. While all views are shown to the student, only the global views are shown to the teacher, thus training a local-to-global correspondence and disentangling objects in feature space.

iBOT~\cite{zhou_ibot_2022} integrates the MIM objective into DINO's framework by masking a randomized selection of input patches from the student's view while still showing them to the teacher. 
Afterwards, the student has to predict the teacher's output for the masked patches.


Finally, AttMask \cite{kakogeorgiou_what_2022} introduces a novel masking strategy that increases the complexity of feature reconstruction compared to the original iBOT masking, aiming to generate a more robust feature space. The final self-attention map generated by the teacher is used to mask the most attended input patches from the student, forcing the student to develop a deeper understanding of the input by masking the most important regions.


\subsection{Local Feature Extraction} 
\label{meth:extract}
We directly use the self-supervised Vision Transformer (ViT) $g$ as a local feature extractor without any additional fine-tuning. A document image \(I\) is cut into \(N\) windows  \( \{w_1, ..., w_N\}\) in a regular grid. The ViT further cuts each window \(w\) into a sequence of flattened patches \(\{p^w_1, ..., p^w_{L}\}\), where each patch \(p^w_i\) is then of length \(P^2\).
A learnable class token [CLS] is prepended, forming the input for the ViT as $ x~=~\{\textsc{[CLS]}, p^w_1, ..., p^w_{L}\}$.
Thus, the output of the ViT $g$ is given as the token sequence 
\begin{equation}
    g(x)~=~\{f^w_\text{[CLS]}, f^w_1, ..., f^w_L\}.
\end{equation}

We find that for aggregating local features using VLAD, retaining the local information of the patch tokens is crucial. However, handwriting images are often sparse due to the horizontal and vertical spacing between words. As a result, many ViT patches may only contain background information, contributing little to the analysis of handwriting characteristics. To address this, we filter out patch tokens that lack sufficient foreground pixels. 
The set of foreground tokens \(FG(I)\) is then given as: 
\begin{equation}
    FG(I) = \{ f^w_i \mid \sum_{j=1}^{P^2} p^w_{i,j} \geq t_{\text{fg}} \text{ for } i \in \{1, \ldots, L\} \text { and } w \in I \},
\end{equation}
where \(p^w_{i,j}\) is the \(j\)-th pixel in the flattened patch and \(t_{\text{fg}}\) a threshold on the number of contained foreground pixels. 

\subsection{Aggregation} \label{meth:aggregate} To construct the VLAD codebook \(\Theta\), we cut all training documents into windows of size 224 x 224 with stride 224, i.e., non-overlapping, and gather all foreground tokens from the entire training set. These are jointly clustered using minibatch $k$-Means~\cite{sculley2010web} with $C$ centroids, which are used as the VLAD~codebook~\( \Theta~=~\{\mu_1, \ldots, \mu_{C}\} \). 
During inference, the test documents are cut into windows with an adjustable stride of \(S_{\text{eval}}\). For each test document \(d\) the set of foreground tokens $FG(d)$ is gathered and encoded by assigning each token \(f \in FG(d)\) to the closest centroid and aggregating the residuals between the centroids and their assigned features.
For a centroid \( \mu_k \), this yields
\begin{equation}
v^d_k = \sum_{\{f \mid \text{NN}_\Theta(f) = \mu_k\}} (f - \mu_k),
\end{equation}
where \(\text{NN}_\Theta(f)\) is the nearest centroid to \(f\) in codebook \(\Theta\).
The resulting VLAD encoding \(\hat v^d\) of a document \(d\) is the concatenation of all such residuals:
\begin{equation}
\hat v^d = \text{concat}(v^d_1, ..., v^d_N).
\end{equation}
We apply power normalization with power 0.5 followed by \(l_2\)-normalization. Finally, principal component analysis (PCA) with whitening is used for decorrelation and dimensionality reduction to \(D\) dimensions, resulting in the global page descriptor~$v^d$ for document $d$. The parameters of the PCA are fitted on the training set.

\subsection{Retrieval and Reranking}\label{meth:rerank}
For retrieval, we use the cosine distance measure. 
A low distance indicates that documents are similar.
The cosine distance \(d_{\text{cos}}\) between two global page descriptors $v^a$ and $v^b$ extracted from documents \(a\) and \(b\) is given as 
\begin{equation}
d_{\text{cos}}(v^a,v^b) = 1 - \frac{v^a \cdot v^b}{||v^a|| \cdot ||v^b||}~.
\end{equation}
We evaluate different reranking strategies from previous methods~\cite{rasoulzadeh_writer_2022,peer_towards_2023} in conjunction with our method, i.e., $k$RNN, Graph reranking and SGR. 

\section{Evaluation Protocol} \label{eval}
In this section, we first introduce the metrics for our evaluation in \Cref{eval:metrics}.
Next, we describe the utilized datasets in \Cref{eval:datasets}.
Lastly, we outline the hyperparameters of our baseline implementation in \Cref{eval:impelemt}. 

\subsection{Metrics} \label{eval:metrics}
The evaluation is done in a leave-one-out fashion, i.e., every document in the test set is used once as a query image. The remaining documents are ranked by their distance to the query such that the documents with the lowest distance rank highest. The relevant documents, i.e., documents written by the same author, should ideally be at the top of this ranking. A common measure to describe the quality of a retrieval list is the mean average precision (mAP).
To assess the writer identification performance, the Top1 accuracy is commonly considered, i.e., the percentage of query images for which the highest ranking result is a relevant document. 

\subsection{Datasets} \label{eval:datasets}
We evaluate our method on two benchmark datasets of historical document images.

\begin{figure}[t]
\centering
\begin{subfigure}{.24\textwidth}
    \centering
    \includegraphics[height=4cm]{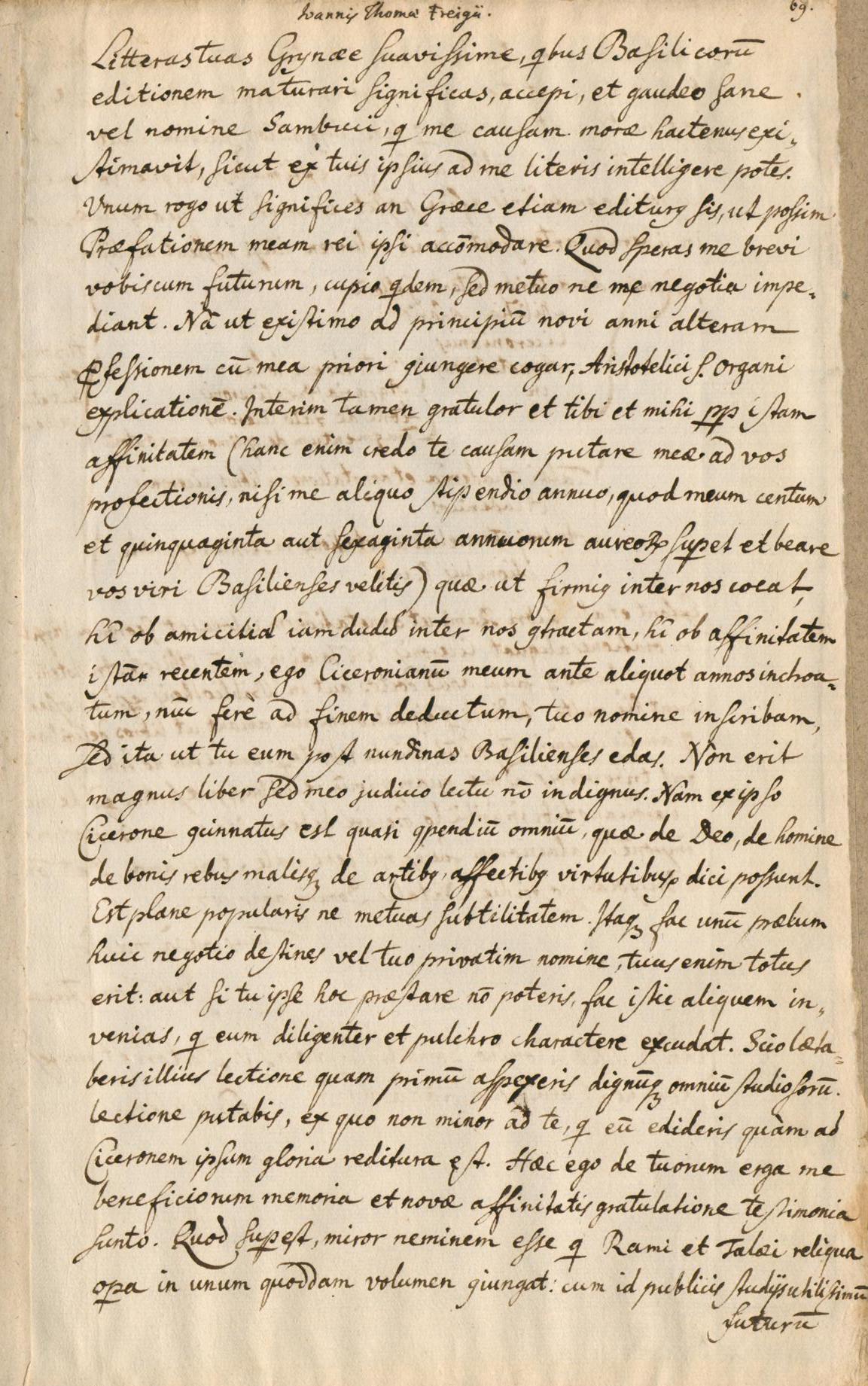}  
    \caption{}
    \label{fig:ic1}
\end{subfigure}
\begin{subfigure}{.5\textwidth}
    \centering
    \includegraphics[height=4cm]{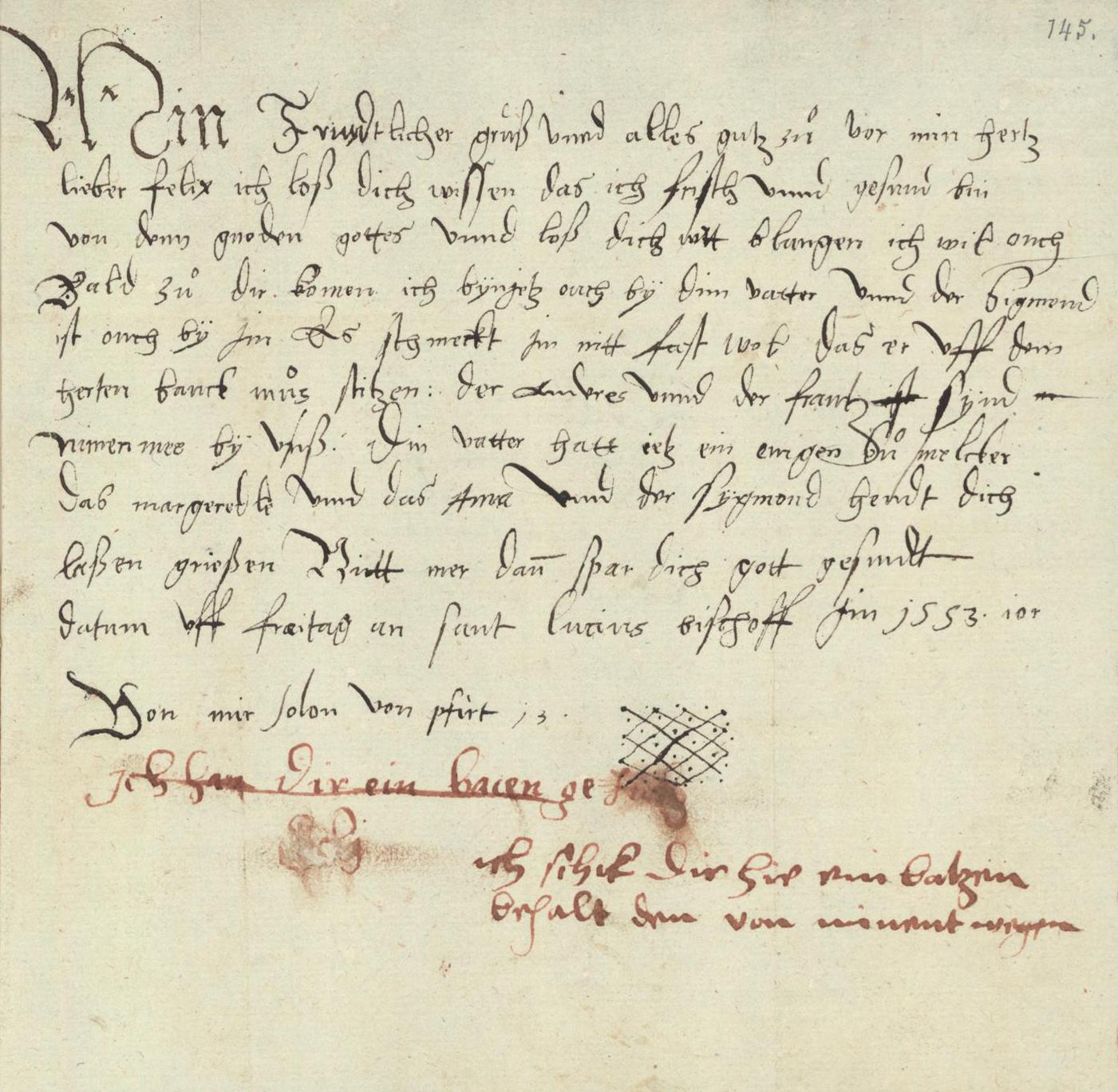}  
    \caption{}
    \label{fig:ic2}
\end{subfigure}
\begin{subfigure}{.24\textwidth}
    \centering
    \includegraphics[height=4cm]{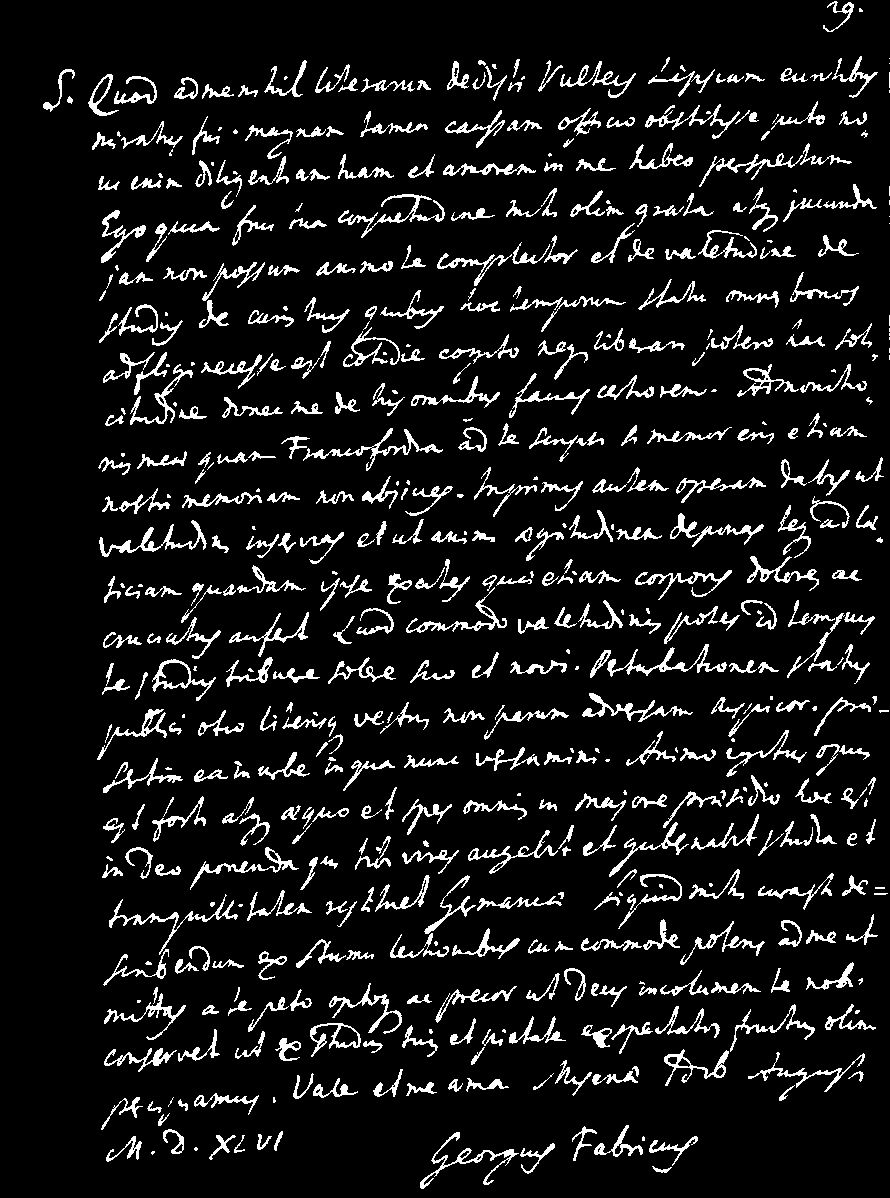}  
    \caption{}
    \label{fig:icbin}
\end{subfigure}
\caption{Visualization of sample document images from the Historical-WI dataset: (a) and (b) show color images, (c) shows a binarized image provided in the dataset.}
\label{fig:icdar}
\end{figure}

\begin{figure}[t]
\centering
\begin{subfigure}{.48\textwidth}
    \centering
    \includegraphics[height=3cm]{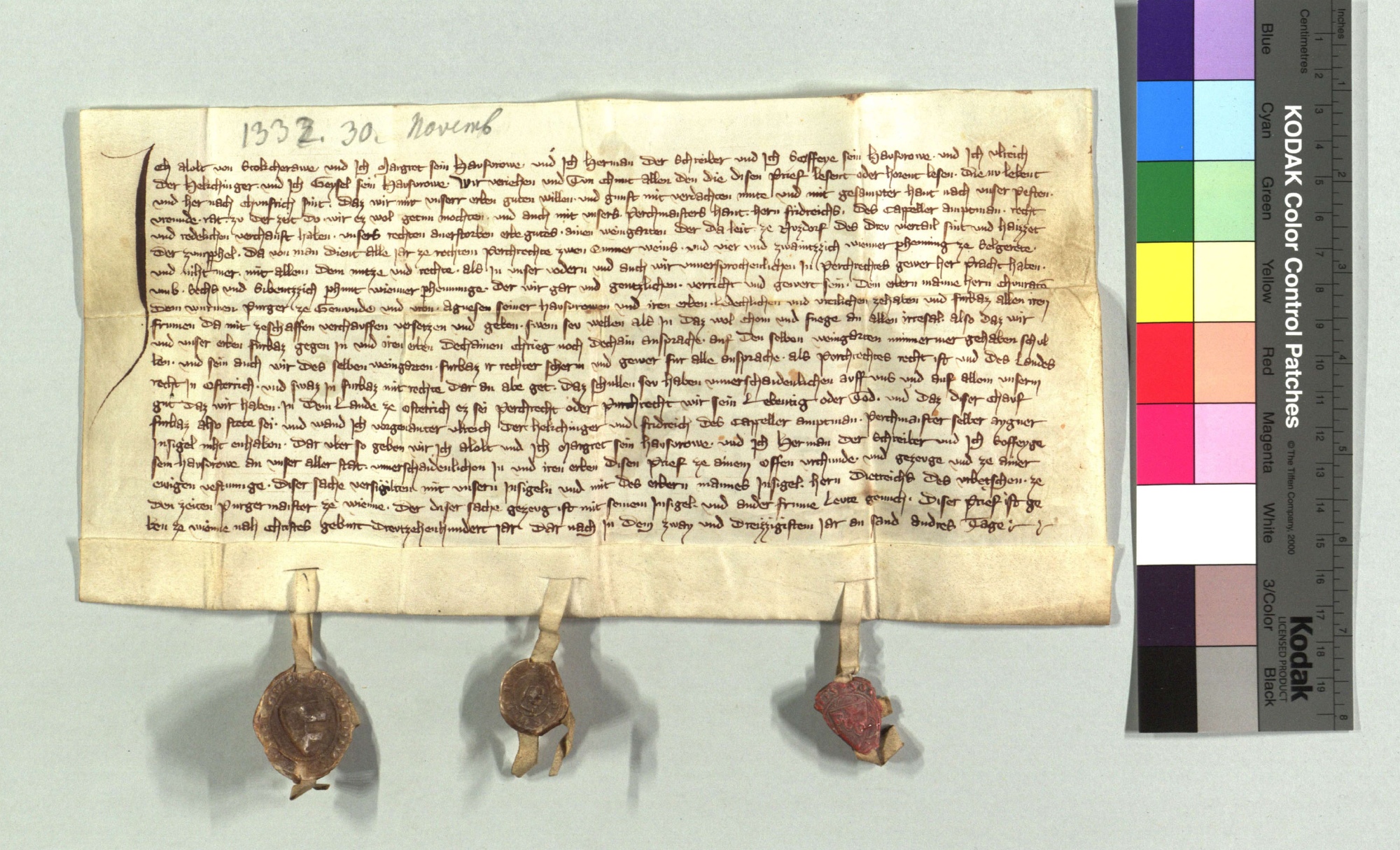}  
\end{subfigure}
\begin{subfigure}{.48\textwidth}
    \centering
    \includegraphics[height=3cm]{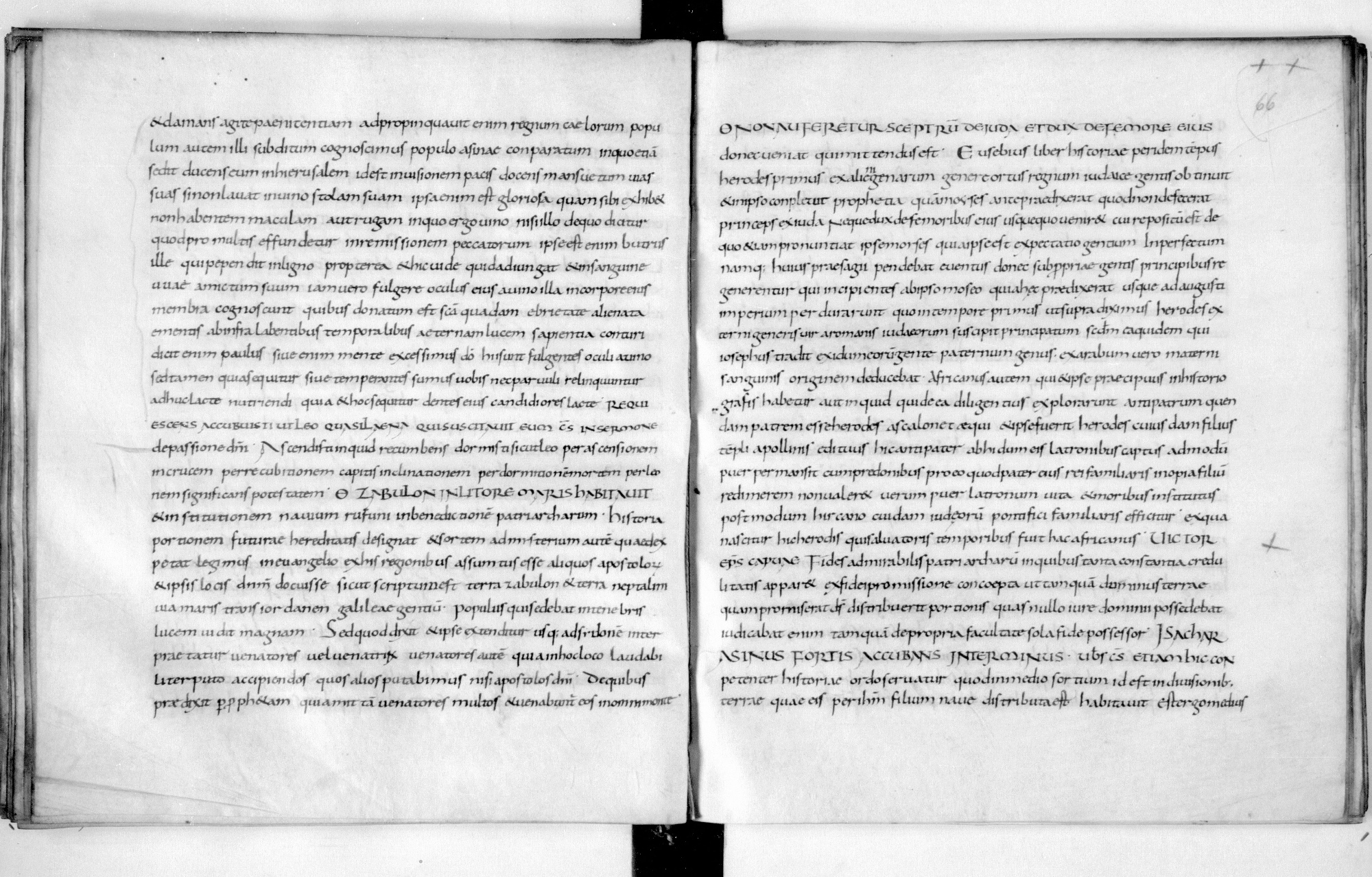}  
\end{subfigure}
\caption{Visualization of sample document images from the HisIR19 dataset.}
\label{fig:hisir}
\end{figure}

\subsubsection{Historical-WI} 
The Historical-WI dataset, a collection of historical document images, was released for the \textit{ICDAR 2017 Competition on Historical Document Writer Identification}~\cite{fiel_icdar2017_2017}. This dataset is available in both binarized and color image formats. It includes a predefined train-test split: the training set comprises 1,182 document images authored by 394 writers, with each writer contributing three documents. The test set is more extensive, containing 3,600 document images from 720 writers, each contributing five documents. Spanning from the 13th to the 20th century, these documents feature texts in German, Latin, and French. \Cref{fig:icdar} shows three examples of contained document images.

\subsubsection{HisIR19}
The HisIR19 dataset was released for the \textit{ICDAR 2019 Competition on Image Retrieval for Historical Handwritten Documents}~\cite{christlein_icdar_2019}. Contrary to the Historical-WI dataset, there is no predefined training split. The authors of the challenge suggest using the Historical-WI test dataset for training. Additionally, a validation dataset is included containing 1200 images of 520 writers, with 300 writers contributing a single page, 100 writers contributing three pages, and 120 writers contributing five pages. The test set is considerably larger than the Historical-WI dataset and contains a total of 20000 documents authored by 10068 writers. 7500 writers contributed one page each, while the others contributed three to five documents. The dataset contains images from manuscript books from the European Middle Ages (9\textit{th} to 15\textit{th} century), letters from the 17\textit{th} and 18\textit{th} centuries, as well as charters and legal documents. \Cref{fig:hisir} shows two examples of contained document images.

\subsection{Implementation Details}\label{eval:impelemt}
We use the following parameters as baseline implementation unless explicitly stated differently. Our model is a ViT-small/16~\cite{dosovitskiy_image_2021} with an input image size of 224. For HisIR19 the document images are only available in color format, thus we binarize them as a preprocessing step using Sauvola Binarization~\cite{sauvola1997adaptive} with a window size of 51. For our evaluations on the HisIR19 dataset, we directly use the ViT trained on the Historical-WI training set and do not perform any additional fine-tuning. We use the HisIR19 validation set only to construct the VLAD codebook.

We generate training data by sampling windows of size 256 in a regular grid with stride 32 from the Historical-WI training set, resulting in roughly 1.75 million training windows. On these, we train the model for 20 epochs. We apply a cosine learning rate schedule with a linear warmup during the first two epochs and a peak learning rate of 0.005. The last layer is frozen during the first epoch.
We use a MultiCrop augmentation~\cite{caron_unsupervised_2021} with two global crops (size 224, scale \( \in [0.4, 1]\)) and  eight local crops (size 96, scale \(\in [0.05, 0.4]\)). 
Since we operate on binary images, all color-related augmentations are dropped.  
Instead, following Peer\textit{ et al.}~\cite{peer_self-supervised_2022},
we apply Dilation and Erosion with random kernels to all crops independently with 50\% probability. 

During feature extraction, we extract windows with stride \(S_{\text{eval}}~=~224\) and apply a foreground threshold of \(t_{\text{fg}}~=~10\) pixels for extracting patch tokens. To save computation, we only use input windows with more than 2.5\% foreground pixels during inference and discard the rest.
For aggregation, we use \(C=100\) centroids in the VLAD codebook and reduce the final dimensionality of our VLAD encodings to \(D=384\) dimensions.

\section{Experiments} \label{exp}
We evaluate the different parts of our method separately. First, in \Cref{exp:extract_aggregate}, we investigate the performance of our foreground tokens in conjunction with different aggregation methods. Second, in \Cref{exp:train}, we compare different training paradigms to train our ViT feature extractor. Third, in \Cref{exp:rerank}, we evaluate different reranking algorithms. Fourth, in \Cref{exp:params}, we investigate the effect of the feature extraction and aggregation parameters. Finally, in \Cref{exp:sota}, we compare our results with previous work.

\subsection{Feature Extraction and Aggregation} \label{exp:extract_aggregate}

\begin{table}[t]
\caption{Evaluation of the performance when extracting class tokens ([CLS]) versus foreground tokens (FG) at different foreground thresholds \(t_\text{fg}\) as features. We evaluate sum-pooling (Sum) and VLAD for aggregation. Results are given for the Historical-WI and HisIR19 test datasets. }\label{tab:extract}
\begin{center}

\begin{tabular}{cc @{\hspace{1em}}cccc @{\hspace{1em}}cccc}
\toprule
& & \multicolumn{4}{c}{Historical-WI} & \multicolumn{4}{c}{HisIR19} \\
\midrule
\multicolumn{2}{c}{} & \multicolumn{2}{c}{Sum} & \multicolumn{2}{c}{VLAD} & \multicolumn{2}{c}{Sum} & \multicolumn{2}{c}{VLAD} \\
Features & $t_{\text{fg}}$ & mAP & Top1 & mAP & Top1 & mAP & Top1 & mAP & Top1 \\
\midrule

[CLS] & - &  \textbf{78.5} & \textbf{90.0} & 64.3 & 80.3 & 92.0& 96.9& 75.8 & 87.8 \\
FG & 0 & 75.1 & 88.3 & 80.1 & 90.5 & 89.8 & 95.8 & 90.1 & 96.0 \\
FG & 1 & 77.1 & \textbf{90.0} & \textbf{81.4} & \textbf{91.1} & 91.5& 96.7& 92.9& 97.2\\
FG & 10 & 76.7& 89.3& 81.1& 90.5& 92.4& \textbf{97.1} & 93.6 & 97.5\\
FG & 20 & 76.3& 89.1& 81.0& 90.6&  \textbf{92.5} & \textbf{97.1} & \textbf{93.7} & \textbf{97.6}\\
FG & 50 & 76.0& 89.3& 80.9& 90.5& 92.3 & \textbf{97.1} & 93.5 & 97.5\\ 
\bottomrule
\end{tabular}
\end{center}
\end{table}

In our proposed method, we extract all foreground patch tokens as features for a given window instead of using the class token. To evaluate this strategy, we compare the performance when using the class tokens as features versus using our foreground patch tokens at various threshold values \(t_\text{fg}\).
For aggregating all local features extracted from a document, we consider sum-pooling and VLAD.

The results given in \cref{tab:extract} show that using class tokens works well with sum-pooling, whereas a significant drop in performance is observed with VLAD. 
In contrast, when using all patch tokens (\(t_\text{fg}=0)\) VLAD outperforms sum-pooling. 
A likely explanation for this is the low number of class features extracted compared to patch tokens.
Filtering empty ViT patches (\(t_\text{fg}=1\)) improves the performance of both encodings compared to using all tokens. Again, VLAD yields better results than sum-pooling. 
Importantly, it also yields better results than sum-pooling of the class tokens on both datasets. 
While further increasing \(t_\text{fg}\) harms performance on the Historical-WI dataset, we observe a peak at \(t_\text{fg}~=~20\) on the HisIR19 dataset. This is likely caused by noise in the automated binarization process which is not present in the curated binarized version of Historical-WI.

\subsection{Vision Transformer Training}
\label{exp:train}
For feature extraction, we train a ViT in a self-supervised approach using AttMask. In this section, we evaluate other self-supervised training approaches, as well as supervised approaches.

\begin{table}[tb]
\caption{
    Evaluation of different training paradigms for training the ViT feature extractor. We evaluated different masking strategies for iBOT and AttMask.
    }\label{tab:train}
\begin{center}

\begin{tabular}{lc @{\hspace{1em}}cccc @{\hspace{1em}}cccc}
\toprule
& & \multicolumn{4}{c}{Historical-WI} & \multicolumn{4}{c}{HisIR19} \\
\midrule
\multicolumn{2}{c}{} & \multicolumn{2}{c}{Sum} & \multicolumn{2}{c}{VLAD} & \multicolumn{2}{c}{Sum} & \multicolumn{2}{c}{VLAD} \\
Method & Features& mAP & Top1 & mAP & Top1 & mAP & Top1 & mAP & Top1 \\
\midrule

Supervised (Writer) &   [CLS] & 52.7 & 72.1 & 27.9 & 40.4 & 82.2 & 90.4 & 53.5 & 69.3 \\
Supervised (Writer) & FG & 67.5 & 84.8 & 61.5 & 79.1 & 87.8 & 94.6 & 82.0 & 91.1\\
Supervised (Page) &     [CLS] & 58.1 & 75.9 & 43.0 & 60.1 & 85.5 & 92.7 & 61.4 & 75.8 \\
Supervised (Page) & FG & 66.4 & 83.5 & 66.8 & 83.4 & 87.8 & 94.7 & 87.0 & 94.2  \\
\midrule

DINO & FG & 74.9 &  89.0 & 80.0 & 90.3& 91.7 & 96.8 & 92.6 & 97.3\\
iBOT (rand) & FG & 75.9 & 88.7 & 80.7 & 90.3& 91.3 & 96.7 & 92.7 & 97.2\\
iBOT (block) & FG & 75.5 & 88.9 & 81.0 & 90.4& 91.2 & 96.5 & 93.2 & 97.3\\
AttMask (Hint) & FG & 75.7 & 88.6 & 81.0 & 90.3& 91.8 & 96.8 & 93.3 & 97.3\\
AttMask (High) & FG & \textbf{76.7} & \textbf{89.3} & \textbf{81.1} & \textbf{90.5} & \textbf{92.4} & \textbf{97.1} & \textbf{93.6} & \textbf{97.5}\\
 \bottomrule
\end{tabular}
\end{center}
\end{table}

\subsubsection{Self-Supervised Training}
In this section, we evaluate other related self-supervised training approaches. 
We compare AttMask \cite{kakogeorgiou_what_2022} to its predecessors, DINO~\cite{caron_unsupervised_2021} and iBOT~\cite{zhou_ibot_2022}, and evaluate different masking strategies.
Both iBOT and AttMask allow to configure the masking process.
Choosing \emph{rand}, the ViT's input patches are masked randomly, whereas when choosing \emph{block} patches for masking are selected, such that they form consecutive block shapes in the original image. 
In the case of AttMask, we evaluate the masking strategies \emph{high} and \emph{hint}.
The masking strategy \emph{high} masks the most highly attended patches in the input image, while \emph{hint} reveals some highly attended patches again. We use the \emph{high} masking strategy as default option in the remaining experiments.
\Cref{tab:train} shows that DINO, iBOT and AttMask slightly improve upon each other.
For all methods, the best results are obtained from encoding our foreground tokens using VLAD, with AttMask achieving slightly higher mAP and Top1 than the others. 

\subsubsection{Supervised Training}
Given the availability of writer identities in our training dataset, a straightforward training approach for the ViT is to use the writer identity as the classification target. We also experiment with using the page id as a classification target. As illustrated in \Cref{tab:train}, both supervised training strategies underperform when compared to self-supervised methods. Interestingly, contrary to our findings in \Cref{exp:extract_aggregate}, the foreground tokens yield better performance with sum-pooling than the class tokens. 

\subsection{Reranking} \label{exp:rerank}

\begin{table}[tb]
\begin{center}
\caption{Evaluation of different reranking methods in combination with class tokens ([CLS]) and foreground tokens (FG), as well as sum-pooling (Sum) and VLAD to compute a global page descriptor.}\label{tab:rerank}
\begin{tabular}{lc @{\hspace{1em}}cccc @{\hspace{1em}}cccc}
\toprule
& & \multicolumn{4}{c}{Historical-WI} & \multicolumn{4}{c}{HisIR19} \\
\midrule
\multicolumn{2}{c}{} & \multicolumn{2}{c}{Sum} & \multicolumn{2}{c}{VLAD} & \multicolumn{2}{c}{Sum} & \multicolumn{2}{c}{VLAD} \\
Method & Features& mAP & Top1 & mAP & Top1 & mAP & Top1 & mAP & Top1 \\
\midrule

None &  [CLS] & {78.5} & \textbf{90.0} & 64.3 & 80.3 & 92.1& 96.9& 79.8& 90.7\\
$k$RNN &  [CLS] & 80.5& 89.0& 66.6& 78.8& \textbf{93.1}& 96.6& 83.1& 90.3\\
Graph & [CLS] & \textbf{80.9}& 89.0& 65.6& 74.8& 93.0& 95.2& 83.2& 88.1\\
SGR &   [CLS] & 80.2& 89.0& 65.5& 75.1& 93.0& 96.2& 81.1& 87.0\\
\midrule
None & FG & 76.7& 89.3& 81.1& \textbf{90.5}& 92.4& \textbf{97.1}& 93.6 & \textbf{97.5} \\
$k$RNN & FG & 78.7& 88.3& \textbf{82.0}& 90.1& \textbf{93.1}& 96.6& \textbf{94.2} & 97.3 \\
Graph & FG & 78.7& 87.3& 81.9& 89.1& \textbf{93.1}& 95.2& 93.9 & 95.6 \\
SGR & FG & 78.2& 87.4& 81.7& 89.4& 92.9& \textbf{97.1}& 93.8 & 96.2\\
\bottomrule
\end{tabular}
\end{center}
\end{table}
In this section, we evaluate the impact of several reranking methods on the performance of our baseline implementation.
We evaluate the kRNN reranking used in~\cite{rasoulzadeh_writer_2022}, Graph reranking~(Graph)~\cite{peer_towards_2023}, and Similarity Graph Reranking (SGR)~\cite{peer_towards_2023}.
We evaluate the impact of reranking in combination with both the class tokens and foreground tokens (\(t_\text{fg}=10 \)), and both sum-pooling and VLAD as encoding. 
To save computation, we don’t optimize the reranking hyperparameters for each combination on the training set but use fixed values which we found to work well across all combinations. For $k$RNN we set \(k=2\). For Graph-reranking we set \(k_1=4, k_2=2, L=3\) following~\cite{peer_towards_2023}. 
For SGR we use \(k=2, \gamma=0.1\). The $\gamma=0.4$ suggested in~\cite{peer_towards_2023} heavily reduced our results, likely due to our higher baseline performance.

\cref{tab:rerank} shows that all reranking methods increase mAP at the cost of Top1 accuracy. On the Historical-WI dataset, sum-pooling of class tokens still produces better results (80.9\% mAP) than foreground tokens (78.7\% mAP). On the HisIR19 dataset, both the class tokens and the foreground tokens achieve equal mAP of 93.1\%, closing the slight gap in the un-reranked results (see \Cref{tab:extract}).
Even with reranking, VLAD computed on the class tokens still heavily underperforms compared to other combinations. The best results overall are still achieved with VLAD on the foreground tokens (82.0\% on Historical-WI, 94.2\% on HisIR19). While all reranking approaches yield similar mAP results, $k$RNN produces slightly better results on both datasets, likely due to retaining the best Top1 accuracy.

\subsection{Parameter Evaluation} \label{exp:params}
In this section, we evaluate the remaining parameters of our pipeline on the Historical-WI dataset. We do not consider encoding class tokens using VLAD as the previous experiments have shown this combination to not yield competitive results.

\subsubsection{Evaluation Stride}
\begin{figure}[tb]  

\begin{minipage}[b]{0.48\textwidth}
\centering
\begin{tikzpicture}
\begin{axis}[
    xlabel={Stride},
    ylabel={mAP [\%]},
    xtick={0, 56, 112, 168, 224},
    x dir=reverse,
    grid=both,
    ymin=70, ymax=85,
    width=\textwidth,
    legend style={nodes={scale=0.8, transform shape}},
    legend pos=south east
]

\addplot[mark=*,green] coordinates {
    (224,78.5) (112,79.2) (56,79.4)
};
\addlegendentry{Sum (CLS)}

\addplot[mark=*,orange] coordinates {
    (224,77.1) (112,77.5) (56,78.2)
};
\addlegendentry{Sum (FG)}


\addplot[mark=*,blue] coordinates {
    (224,81.4) (112,82.0) (56,82.6)
};
\addlegendentry{VLAD (FG)}
\end{axis}
\end{tikzpicture}
\end{minipage}
\hfill
\begin{minipage}[b]{0.48\textwidth}
\centering
\begin{tikzpicture}
\begin{axis}[
    xlabel={Stride},
    ylabel={Top1 Accuracy [\%]},
    xtick={0, 56, 112, 168, 224},
    x dir=reverse,
    grid=both,
    ymin=85, ymax=95,
    width=\textwidth,
    legend style={nodes={scale=0.8, transform shape}},
    legend pos=south east
]

\addplot[mark=*,green] coordinates {
    (224,90.0) (112,90.2) (56,90.6)
};
\addlegendentry{Sum (CLS)}

\addplot[mark=*,orange] coordinates {
    (224,90.0) (112,90.0) (56,90.3)
};
\addlegendentry{Sum (FG)}


\addplot[mark=*,blue] coordinates {
    (224,91.1) (112,91.3) (56,91.9)
};
\addlegendentry{VLAD (FG)}

\end{axis}
\end{tikzpicture}
\end{minipage}

\caption{Evaluation of \(S_\text{eval}\),i.e., the stride with which windows are sampled from the test documents during inference on the Historical-WI dataset. We compare different combinations of features and aggregating methods. The left plot shows mAP and the right plot shows Top1 accuracy. }
\label{fig:stride}

\end{figure}
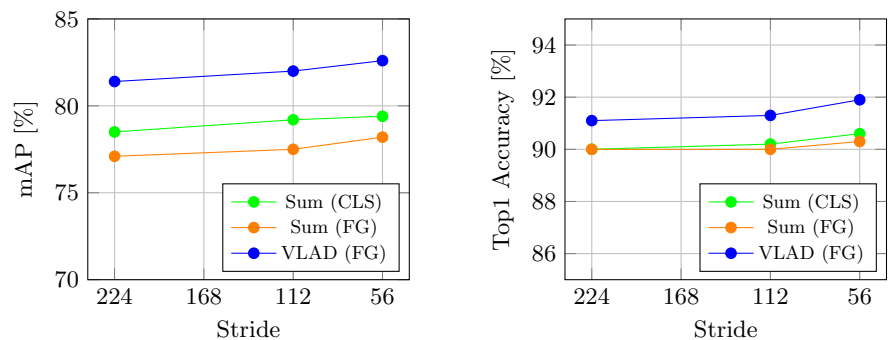
During inference, we sample windows in a regular grid with stride \(S_\text{eval}\). In the previous experiments, we used a baseline value of \(S_\text{eval}=224\). \Cref{fig:stride} shows that reducing the stride enhances performance in all cases. Lowering \(S_\text{eval}=224\) to 56 improves the performance of VLAD on the foreground tokens to 82.6\% mAP (+1.5\%). We did not evaluate smaller strides for computation reasons as halving the stride produces four times more input windows. 

\subsubsection{Number of VLAD Cluster Centers}
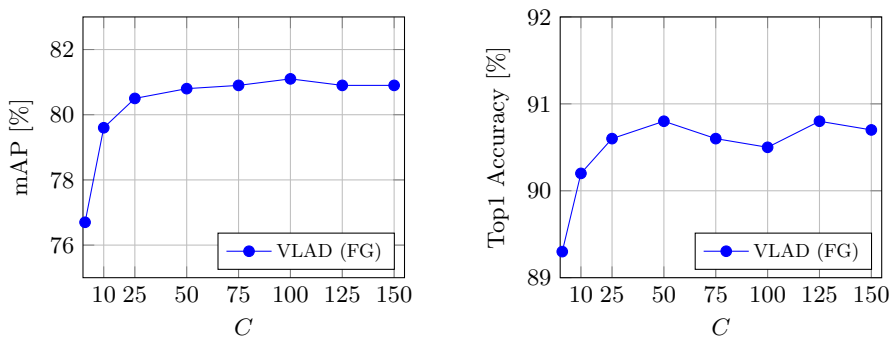
\begin{figure}[tb]  

\begin{minipage}[b]{0.48\textwidth}
\centering
\begin{tikzpicture}
\begin{axis}[
    xlabel={$C$},
    ylabel={mAP [\%]},
    xmin=0, xmax=155,
    xtick={10, 25, 50, 75, 100, 125, 150},
    grid=both,
    ymin=75, ymax=83,
    width=\textwidth,
    legend style={nodes={scale=0.8, transform shape}},
    legend pos=south east
]


\addplot[mark=*,blue] coordinates {
    (1, 76.7) (10, 79.6) (25,80.5) (50,80.8) (75,80.9) (100,81.1) (125, 80.9) (150, 80.9)
};
\addlegendentry{VLAD (FG)}
\end{axis}
\end{tikzpicture}
\end{minipage}
\hfill
\begin{minipage}[b]{0.48\textwidth}
\centering
\begin{tikzpicture}
\begin{axis}[
    xlabel={$C$},
    xmin=0, xmax=155,
    ylabel={Top1 Accuracy [\%]},
    xtick={10, 25, 50, 75, 100, 125, 150},
    grid=both,
    ymin=89, ymax=92,
    width=\textwidth,
    legend style={nodes={scale=0.8, transform shape}},
    legend pos=south east
]


\addplot[mark=*,blue] coordinates {
    (1, 89.3) (10, 90.2) (25,90.6) (50,90.8) (75,90.6) (100,90.5) (125, 90.8) (150, 90.7) 
};
\addlegendentry{VLAD (FG)}
\end{axis}
\end{tikzpicture}
\end{minipage}

\caption{Evaluation of parameter $C$, i.e., the number of cluster centers used to compute the VLAD codebook $\Theta$. The left plot shows mAP and the right plot shows Top1 accuracy.}
\label{fig:clusters}

\end{figure}

Our baseline constructs a codebook of size \(C=100\) for the VLAD encoding. \Cref{fig:clusters} shows that both mAP and Top1 are relatively stable regardless of the number of clusters. Even with as few as 10 clusters performance only deteriorates slightly, and still considerably improves on sum-pooling. 

\subsubsection{PCA dimensionality}
\begin{figure}[tb]  

\begin{minipage}[b]{0.48\textwidth}
\centering
\begin{tikzpicture}
\begin{axis}[
    xlabel={$D$},
    ylabel={mAP [\%]},
    xmin=128, xmax=1050,
    xtick={128, 256, 384, 512, 768, 1024},
    grid=both,
    ymin=75, ymax=85,
    width=\textwidth,
    legend style={nodes={scale=0.8, transform shape}},
    legend pos=south east
]

\addplot[mark=*,green] coordinates {
    
    (128, 78.4) (256, 78.9) (384, 78.2)
};
\addlegendentry{Sum (CLS)}

\addplot[mark=*,orange] coordinates {
    (128,75.5) (256, 77.0) (384, 76.5)
};
\addlegendentry{Sum (FG)}

\addplot[mark=*,blue] coordinates {
    (128,78.9) (256, 80.6) (384, 81.1) (512, 81.0) (786, 80.6) (1024, 79.9)
};

\addlegendentry{VLAD (FG)}
\end{axis}
\end{tikzpicture}
\end{minipage}
\hfill
\begin{minipage}[b]{0.48\textwidth}
\centering
\begin{tikzpicture}
\begin{axis}[
    xlabel={$D$},
    ylabel={Top1 Accuracy [\%]},
    xmin=128, xmax=1050,
    xtick={128, 256, 384, 512, 768, 1024},
    grid=both,
    ymin=85, ymax=95,
    width=\textwidth,
    legend style={nodes={scale=0.8, transform shape}},
    legend pos=south east
]

\addplot[mark=*,green] coordinates {
    (128, 89.6) (256, 89.9) (384, 90.0)
};
\addlegendentry{Sum (CLS)}

\addplot[mark=*,orange] coordinates {
    (128,88.3) (256, 89.2) (384, 89.0) 
};
\addlegendentry{Sum (FG)}


\addplot[mark=*,blue] coordinates {
    (128,89.7) (256, 90.4) (384, 90.7) (512, 91.1) (786, 91.0) (1024, 90.9)
};
\addlegendentry{VLAD (FG)}

\end{axis}
\end{tikzpicture}
\end{minipage}

\caption{Evaluation of the parameter $D$, i.e. the number of dimensions kept during principal component analysis. We evaluate different combinations of features and aggregation methods. The left plot shows the mAP and the right plot shows the Top1 accuracy.}
\label{fig:dim}

\end{figure}
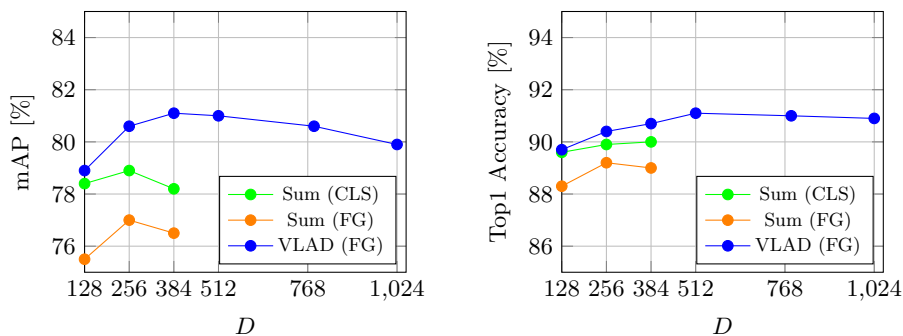

After aggregation, we use principal component analysis with whitening and dimensionality reduction to \(D=384\) dimensions. As \Cref{fig:dim} shows, retrieval performance with VLAD peaks at $D=384$ dimensions, whereas Top1 accuracy peaks at $D=512$. With sum-pooling, the dimensionality of the final page descriptor is equal to the ViT's embedding dimensionality, in our case 384. As such, larger values can not be evaluated. For both class tokens and foreground tokens, mAP and Top1 peak around 256 dimensions but still fall short compared to VLAD.

\subsection{Comparison with State-of-the-Art} \label{exp:sota}

For our comparison with other methods, we distinguish between the performance without additional reranking steps and the performance when reranking is applied. We use the baseline parameters outlined in \cref{eval:impelemt}, with the exception of reducing the evaluation stride, i.e., setting \(S_\text{eval}=56\). The results are given in \Cref{tab:sota}.

\subsubsection{Historical-WI}
On the Historical-WI dataset, our method surpasses existing methods considerably in terms of mAP. We achieve 82.6\% mAP and 91.9\% Top1 score without reranking, beating the best previous method of Lai\textit{ et al.} (77.1\% mAP, 90.1\% mAP) by 5.5\% mAP and 0.8\% Top1 score. Notably, our method also exceeds previous methods when using sum-pooling as an encoding in conjunction with the ViTs class token. We achieve 79.4\% mAP with this configuration, beating previous methods by more than 2\% mAP. This is especially noteworthy as methods based on CNN-based features perform much worse with sum-pooling: Christlein\textit{ et al.}\cite{christlein_unsupervised_2017} report a drop of over 30\% mAP compared to their $m$VLAD encoding. 
When applying additional reranking, our method still beats previous methods. We achieve 83.1\% mAP with reranking, improving over the method of Peer\textit{ et al.} (80.6\% mAP)~\cite{peer_towards_2023} by 2.5\% mAP. Still, even the sum-pooling of class tokens outperforms previous methods slightly. 

\subsubsection{HisIR19}
On the HisIR19 dataset, our method also outperforms previous methods considerably. Without reranking, we achieve 94.4\% mAP and 97.8\% Top1 accuracy, beating the previously best method (92.8\% mAP, 97.4\% Top1\cite{lai_encoding_2020}) by 1.6\% mAP and 0.4\% Top1 accuracy. When also applying reranking, our method achieves 95.0\% mAP and 97.6\% Top1 accuracy, improving over the best previous method by 1.8\% mAP and 0.9\% Top1 accuracy. Similar to our findings for the Historical-WI dataset, the sum-pooled class tokens achieve competitive performance, with and without reranking. 

\subsubsection{CVL}
Additionally, we evaluate our method on the CVL database~\cite{kleber2013cvl}, a dataset containing modern handwriting. We directly use the ViT feature extractor trained on the Historical-WI dataset without any fine-tuning and construct the VLAD codebook from the training set of the CVL dataset. With reranking, we achieve 98.6\% mAP and 99.4\% Top1 accuracy, matching the results of previous methods \cite{8395190,rasoulzadeh_writer_2022}. Even without reranking, our method achieve a mAP of 97.1\%, highlighting the robustness of the extracted features, despite only training on a relatively small set of historical documents.

\begin{table}[tb]
\caption{
Comparison of our method with the state of the art on the Historical-WI, HisIR19 and CVL test datasets. We evaluate two configurations of our method: aggregating class tokens with sum-pooling and aggregating foreground tokens with VLAD. For all datasets, we use the same evaluation parameters, i.e., \(D=384, C=100, t_\text{fg}=10, k=2, S_{eval}=56\). } \label{tab:sota}
\begin{center}

\begin{tabular}{l cc c@{\hspace{0.5em}}c@{\hspace{1em}}c@{\hspace{0.5em}}c@{\hspace{1em}}c@{\hspace{0.5em}}c@{\hspace{1em}}c@{\hspace{0.5em}} c@{\hspace{1em}} c c }
\toprule
 && & \multicolumn{2}{c}{Historical-WI} & \multicolumn{2}{c}{HisIR19} & \multicolumn{2}{c}{CVL} \\
Method & Encoding & Reranking & mAP & Top1 & mAP & Top1 & mAP & Top1\\
\midrule
Peer\textit{ et al.} \cite{peer_towards_2023} & NetVLAD & - & 73.4 & 88.5 & 91.6 & 96.1&- &- \\
Christlein\textit{ et al.} \cite{christlein_unsupervised_2017} & mVLAD & - & 74.8 & 88.6 & - & - &- &- \\  
Lai\textit{ et al.} \cite{lai_encoding_2020} & bVLAD & - & 77.1 & 90.1 & 92.5 & {97.4}&- &- \\
\textbf{Ours (CLS)} & Sum & - & 79.4 & 90.6 & 92.8 & 97.3& 94.2 & 98.9 \\
\textbf{Ours (FG)} & VLAD & - & \textbf{82.6} & \textbf{91.9} & \textbf{94.4} & \textbf{97.8}& 97.1 & 99.4 \\
\midrule
Christlein\textit{ et al.} \cite{8395190} & VLAD & E-SVM &- &- &- &- & 98.4 & \textbf{99.5}  \\
Rasoulzadeh \cite{rasoulzadeh_writer_2022} & NetVLAD & $k$RNN &- &- &- &- & \textbf{98.6} & 99.2  \\
Christlein\textit{ et al.} \cite{christlein_unsupervised_2017} & mVLAD & E-SVM & 76.2 & 88.7 & 91.2 & 97.0 \vspace{0.0cm} &- &-\\
Christlein\textit{ et al.} \cite{christlein_unsupervised_2017}& mVLAD & P/T-SVM \cite{jordan_re-ranking_2020} & 78.2 & 89.4 & - & - \vspace{0.0cm} &- &-\\
Peer\textit{ et al.} \cite{peer_towards_2023} & NetVLAD & SGR & 80.6 & \textbf{91.1} & 93.2 & 96.7 \vspace{0.0cm}&- &-\\
\textbf{Ours (CLS)} & Sum & $k$RNN & 81.2 & 90.6 & 93.2 & 96.9& 97.6 & 98.8 \\
\textbf{Ours (FG)} & VLAD & $k$RNN & \textbf{83.1} & {90.9} & \textbf{95.0} & \textbf{97.6} & \textbf{98.6} & 99.4  \\ 
\bottomrule
\end{tabular}
\end{center}
\end{table}

\section{Conclusion} \label{conc}
In this work, we presented a novel method using a Vision Transformer as a feature extractor. The model is trained in an unsupervised fashion. Patch tokens containing foreground are extracted as local features and subsequently encoded with VLAD. Retrieval is done using the cosine distance, with optional reranking. Our method achieved a new state-of-the-art performance on the historical benchmark datasets Historical-WI and HisIR19, improving over previous methods by 2.5\% mAP and 1.8\% mAP respectively. We additionally showed that our method is versatile and also works well on modern datasets, achieving 98.6\% mAP on the CVL database without requiring any fine-tuning of the ViT. 

Further research could be done to evaluate other SSL methods for model training and different model architectures. In terms of SSL methods, DINOv2~\cite{oquab2023dinov2} introduced several improvements to the iBOT framework which might be interesting for writer retrieval, e.g. the KoLeo regularizer~\cite{sablayrolles2018spreading}. Regarding architectures, Swin-Transformers~\cite{liu2021swin} have shown promising results in domains with limited data, which might help to boost performance and training time. 

Moreover, we showed that only considering patch tokens containing sufficient foreground information is beneficial. Here, future research could investigate other strategies for filtering out patch tokens, for instance by utilizing the self-attention of the ViT to identify relevant patches.

\bibliographystyle{splncs04}
\bibliography{bib}





\end{document}